\documentclass[5p]{elsarticle}
\usepackage{prletters}

\usepackage{url}
\usepackage{todonotes}
\usepackage{amsmath}
\usepackage{amssymb}
\usepackage{physics}
\usepackage{booktabs}
\usepackage{float}
\usepackage{comment}
\usepackage{arydshln}

\usepackage{graphicx}
\usepackage[normalem]{ulem}
\useunder{\uline}{\ul}{}
\usepackage[highlight]{verlab}
\usepackage[english]{babel}
\usepackage[T1]{fontenc}

\usepackage{url}
\usepackage{amssymb}
\usepackage{hyperref}
\usepackage[figuresright]{rotating}

\begin{document}

\begin{frontmatter}

%% Title, authors and addresses

%% use the tnoteref command within \title for footnotes;
%% use the tnotetext command for the associated footnote;
%% use the fnref command within \author or \address for footnotes;
%% use the fntext command for the associated footnote;
%% use the corref command within \author for corresponding author footnotes;
%% use the cortext command for the associated footnote;
%% use the ead command for the email address,
%% and the form \ead[url] for the home page:
%%
% \title{Title\tnoteref{label1}}
% \tnotetext[label1]{}
% \author{Name\corref{cor1}\fnref{label2}}
% \ead{email address}
% \ead[url]{home page}
% \fntext[label2]{}
% \cortext[cor1]{}
% \address{Address\fnref{label3}}
% \fntext[label3]{}

% \dochead{}
%% Use \dochead if there is an article header, e.g. \dochead{Short communication}

\title{Improving the matching of deformable objects by learning to detect keypoints}

%\title{On the improvement of matching deformable objects by learning to detect keypoints}
%\title{The Deformable Imitation Game: Learning to Detect by Imitating Good Matches Affected by Deformations}
%\title{Learning to Detect Visual Keypoints to Match Non-Rigid Objects}
%\title{Learning from Base Detectors: A Novel Keypoint Detection Method for Non-Rigid Image Correspondence}

%% use optional labels to link authors explicitly to addresses:
\author[ufmg]{Felipe Cadar \corref{cor1}}
\ead{cadar@dcc.ufmg.br}

\author[ufmg]{Welerson Melo}
\ead{welerson.melo@dcc.ufmg.br}

\author[ubfc]{Vaishnavi Kanagasabapathi}
\ead{Vaishnavi_Kanagasabapathi@etu.u-bourgogne.fr}

\author[ufmg]{Guilherme Potje}
\ead{guipotje@dcc.ufmg.br}

\author[ubfc,inria]{Renato Martins}
\ead{renato.martins@u-bourgogne.fr}

\author[ufmg]{Erickson R. Nascimento}
\ead{erickson@dcc.ufmg.br}

\address[ufmg]{Department of Computer Science, Universidade Federal de Minas Gerais, Brazil}
\address[ubfc]{Laboratoire Interdisciplinaire Carnot de Bourgogne, UMR 6303 CNRS, Université de Bourgogne, France}
\address[inria]{TANGRAM team, LORIA, Inria, Université de Lorraine, France}
\cortext[cor1]{Corresponding Author}

\begin{abstract}
We propose a novel learned keypoint detection method to increase the number of correct matches for the task of non-rigid image correspondence. By leveraging true correspondences acquired by matching annotated image pairs with a specified descriptor extractor, we train an end-to-end convolutional neural network (CNN) to find keypoint locations that are more appropriate to the considered descriptor. %For that, we apply geometric and photometric warpings to images to generate a supervisory signal, allowing the optimization of the detector. 
Experiments demonstrate that our method enhances the Mean Matching Accuracy of numerous descriptors when used in conjunction with our detection method, while outperforming the state-of-the-art keypoint detectors on real images of non-rigid objects by $20$ p.p. We also apply our method on the complex real-world task of object retrieval where our detector performs on par with the finest keypoint detectors currently available for this task. The source code and trained models are publicly available at \url{https://github.com/verlab/LearningToDetect_PRL_2023}

\end{abstract}


\end{frontmatter}

%%
%% Start line numbering here if you want
%%
% \linenumbers

%% main text
\section{Introduction}
\label{sec:intro}
Finding discriminative patterns that are repeatable across various images of the same scene is the primary goal of keypoint detectors. A good detector should be equivariant to changes in viewpoint and scale as well as invariant to changes in illumination. Furthermore, objects may change in shape over time due to deformations. Therefore, equivariance to non-rigid deformations is an important property when identifying points for visual correspondence. 

Recently, learning-based systems~\cite{di2018kcnn, laguna2022keynet, revaud2019r2d2, dusmanu2019d2net, luo2020aslfeat, tyszkiewicz2020disk, song2020sekd, suwanwimolkul2021learning,vasconcelos2017kvd} have become more prevalent in feature detection techniques, producing outcomes that greatly exceed handcrafted keypoint detectors~\cite{lowe2004sift, rublee2011orb}. 
Despite the advances in learning feature representations, end-to-end learning of keypoint detection is still a challenging problem. Typically, a learned keypoint detector is trained to find recurring patterns across images. However, this approach does not ensure that the detected points are good for matching. 
For instance, consider the high degree of ambiguity in texture while matching points on edges and repetitive patterns, \eg, structures that commonly appear in man-made buildings. In regions where accurate matching is not achievable~\cite{revaud2019r2d2}, methods that first detect keypoints before describing them often degrade the matching performance. Moreover, it is computationally infeasible to bypass detection by dense matching, i.e., matching all pixels~\cite{tyszkiewicz2020disk}. The time complexity to match two images $A$ and $B$ using feature sets $F_A$ and $F_B$ is $O(|F_A|\cdot|F_B|)$. The problem becomes intractable, requiring carefully designed training techniques and high computational resources. Dense matching also has to deal with the inherent ambiguity caused by poorly-textured regions and poor localization properties. Some methods, like LoFTR~\cite{sun2021loftr}, rely on fewer confident matches at a lower resolution and propagate this more robust information of matching for neighboring pixels to higher resolutions. In this context, keypoint detection may be needed to improve  dense matching. Few works have attempted to achieve equivariance regarding non-rigid deformations, and most recent advances in this direction, such as the work of Yu~\etal~\cite{you2020ukpgan}, suggest using additional depth information. Still, color cameras are by far the most used and available imaging sensors. 

Differently from these previous approaches, we propose a novel learned keypoint detection methodology for handling non-rigid deformations on still images (Figure~\ref{fig:qualitative} shows some qualitative results). In this work, we assume locally smooth non-rigid deformations, i.e., the gradient of the deformation field is locally continuous and does not have abrupt discontinuities. This implies that the nearby pixels from a point in the image should follow similar offsets from one image to another, which is a reasonable assumption for most real-world scenarios, especially when dealing with local features, because of their local receptive field. Inspired by DEAL \cite{potje2021extracting}, we model the deformations using Thin-plate Splines \cite{BartoliPC10}, where one minimizes the bending energy of the deformation field by solving for an affine transformation and the Radial Basis Functions (RBFs) coefficients for the following energy function: 

\begin{equation}
    E_{tps}(f) = \sum_{i=1}^{k}||\mathbf{y_i} - f(\mathbf{x_i})||^2,
    \label{eq:tps1}
\end{equation}
\noindent where $\mathbf{y_i}$ and $\mathbf{x_i}$ are two sets of corresponding points, also called control points, and $f(\mathbf{x})$ is the RBF that employs the TPS kernel. Once the affine matrix $\mathbf{A}$ and coefficients of $\mathbf{w_i} \in \mathbb{R}^2$ are obtained, one can map a pixel $\mathbf{x}$ from a reference frame to the target frame by using the TPS warp as follows:
\begin{equation}
    f(\mathbf{x}) = \mathbf{Ax} + \sum_{i=1}^{n_c}\rho(||\mathbf{x} - \mathbf{c_i}||^2)\mathbf{w_i},
     \label{eq:tps2}
\end{equation}
\noindent where $\rho(r) = r^2 \ log \ r$ is the thin-plate radial basis function, $\mathbf{x}$ is a 2D point, $n_c$ is the number of control points, $\mathbf{c_i}$ is a control point of index $i$.

Several challenges need to be addressed for successfully detecting keypoints on deformable surfaces with high accuracy. Specifically, robustness to non-linear illumination changes, occlusions, and local photometric changes, which can introduce high ambiguity for both the detection and description stages. By assuming that desirable features to be discovered are also salient points that are likely to produce accurate matches, we propose to tackle the keypoint detection problem in a well-defined manner. By using an existing detector-descriptor setup, our learned detection model takes advantage of matching score maps. Our network can be easily paired with any combination of pre-existing detector-descriptor since it is trained to identify good features based on the map created from true descriptor matches on virtual deformations. 

We test our detector using three separate benchmark datasets of real deformable objects as well as with an application of content-based object retrieval, demonstrating that our approach can achieve state-of-the-art performance not only in matching evaluation scores but also in a relevant practical computer vision task of image retrieval. Figure~\ref{fig:qualitative} shows how well our detector performs in matching keypoints in comparison to several recently developed detectors.

The two main technical contributions of our work are: (i) a novel keypoint detection training framework designed to enhance the matching performance of pre-existing descriptors and (ii) the first learned keypoint detector tailored to handle non-rigid deformations on color images. 
This paper extends the results and contributions of our previous work~\cite{cit:welerson2022} on several major fronts. First, we introduce and evaluate novel architecture components, such as deformable convolutional network (DCN) layers, to evaluate the contribution of providing the network with more specialized inductive bias than regular CNNs for the task of non-rigid keypoint detection. Second, we perform experiments by adding DCN layers to different stages of the network backbone. Extensive experimental analysis reveals that the DCN layers help during training but do not improve the generalization of the network; the main improvement comes from our learning strategy. Finally, we extend the experimental evaluation with nine more recent strong baseline detectors. These additional experiments with several keypoint detectors indicate that our novel training paradigm can be adopted as a general framework for improving matching for non-rigid correspondence. Our method can be trained for any pair of detectors and descriptors to specialize the keypoint selection and improve matching quality with a specified descriptor.

\section{Related work}
\label{sec:related}
The most popular keypoint detectors are based on carefully crafted algorithms that choose feature patterns like blobs, corners, or edges~\cite{harris1988combined, lowe2004sift, rublee2011orb}. Deep learning, on the other hand, has recently emerged as the new standard for feature extraction and image matching. However, these learning-based methods are mainly used for tasks involving feature description~\cite{luo2019contextdesc, wu2020aggregation, potje2021extracting}, and joint detection and description~\cite{ revaud2019r2d2, dusmanu2019d2net, luo2020aslfeat, song2020sekd, tyszkiewicz2020disk, detone2018superpoint, ono2018lfnet, yi2016lift, Zhao2022ALIKE}. There are only a few studies that specifically address learning to recognize keypoints by raising detection likelihood for repeatable areas between image pairs~\cite{savinov2017quad, zhang2018learning, Verdie2015tilde}. While such keypoint detectors have high repeatability, their keypoints are mostly ambiguous, and the matching performance is degraded \cite{laguna2022keynet}. To enhance the representation and generalization of low-level characteristics, Barroso-Laguna~\etal~\cite{laguna2022keynet} and Di~\etal~\cite{di2018kcnn} employ fixed handcrafted filters in addition to learned ones. However, manually created filters can include biases that could make it difficult to find appropriate keypoints, resulting in subpar detections. 

Alternatively, recent studies have been published based on a new research trend called the describe-then-detect approach, where the keypoints are found from a dense descriptor map. According to Revaud~\etal~\cite{revaud2019r2d2}, keypoints should be identified based on repeatability and dependability because detection and description are inextricably entwined. Suwanwimolkul~\etal~\cite{suwanwimolkul2021learning} noted that since the keypoint selection is more handcrafted in approaches like D2-Net~\cite{dusmanu2019d2net} and ASLFeat~\cite{luo2020aslfeat}, there are no guarantees that the chosen keypoints can match the learned descriptors. Consequently, the accuracy of the matched keypoints is not usually very high. In contrast, in our method, the detector is trained to increase matching accuracy with real image pairings, and the peaks are generated directly from the network output (score map shown in Figure~\ref{fig:architecture}).

Only a few studies take descriptor matching into account in the training pipeline. GLAM~\cite{truong2019glampoints} detects keypoints based on matching quality but for a very specific domain of retinal images. By first identifying repeatable keypoints and then selecting the confident keypoints based on the matching, SEKD~\cite{song2020sekd} proposes a non-domain specific detector and descriptor. However, when a high number of good keypoints for matching is not discovered during the repeatability optimization stage, this strategy often produces degraded results. To jointly optimize detection and description, DISK~\cite{tyszkiewicz2020disk} takes into account detection and description in a probabilistic relaxation and applies a reinforcement learning strategy. The method's disadvantage is that in order to converge, it needs cautious hyperparameter adjustments and parameter annealing tricks. A decision tree was developed by Tonioni~\etal~\cite{tonioni2018learning} to learn how to choose 3D keypoints based on good matches. The authors argue that good traits for detection are those that are most likely to produce accurate matches. We implement a similar approach but on 2D keypoints, using a modern deep learning design. We employ the outcomes of matching descriptors with a weighting technique for creating a Matching Heatmap, which is non-domain specific. 

Descriptors like DEAL~\cite{potje2021extracting} present a deformation-aware local feature description technique that learns to describe non-rigid surfaces without depth information using explicit network deformation modeling. Similarly, the DaLi descriptor~\cite{moreno2011deformation} finds features invariant to non-rigid deformations and variations in light by transforming local patches into meshes and applying heat diffusion operators. Likewise, GeoBit descriptor~\cite{nascimento2019geobit} and GeoPatch~\cite{potje2022cviu} compute isometrically invariant features for RGB-D images by using geodesics from an object's surface. In contrast to the suggested strategy in this paper, these solutions are designed for RGB-D images and focus only on the description step and ignore the detection phase. A 3D keypoint detector has recently been proposed by UKPGAN~\cite{you2020ukpgan}, in which keypoints are found for 3D reconstruction. However, unlike our detector for 2D images, their approach is only appropriate in the context of 3D keypoints. To the extent of our knowledge, no previous study has addressed the detection of 2D features on images with non-rigid deformations. Tarashima \etal \cite{Tarashima2018} proposed to deal with non-rigid objects by considering deformations only in the matching stage. Despite having reasonable premisses for matching, the deformations still degrade the detection and description needed as input in the matching.
In this paper, we offer a method for obtaining keypoints less affected by non-rigid deformations using only visual data.
\section{Methodology}
\label{sec:methodology}

\begin{figure}[!t]
%\vspace{-20pt}
  \centering
    \includegraphics[width=.9\linewidth]{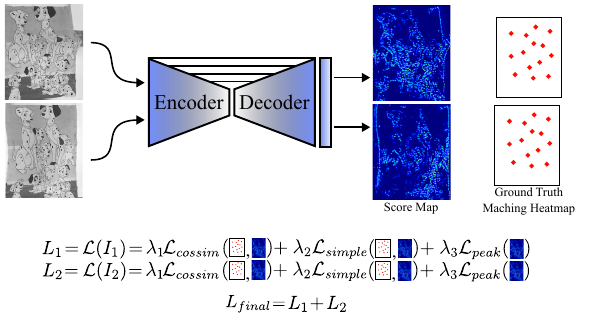}
    \hfill
    \vspace*{-0.5cm}
    \caption{\textbf{Network architecture for keypoint detection.} The Siamese network is optimized to detect reliable keypoints to be matched for a given descriptor. The encoder-decoder architecture is an U-Net with dimensions $D_{encoder}=\{1, 32, 64, 128\}$ and $D_{decoder}=\{128, 64, 32, 1\}$ with skip connections and a final sigmoid layer.}
    \label{fig:architecture}
    %\vspace{-15pt}
\end{figure}

Our detector is trained to identify stable areas in images that are affected by non-rigid deformations. Our training setting imposes keypoints to be found at repetitive places with a high likelihood of matching for a given descriptor. Our network (Figure~\ref{fig:architecture}) receives an image and outputs the most likely keypoints to be detected. We describe the training framework, the loss design, and implementation specifics in the following sections.

\subsection{Keypoint detection learning framework}
\label{subsec:framwork}
In contrast to learned keypoints like Key.net~\cite{laguna2022keynet} and describe-to-detect techniques like R2D2~\cite{revaud2019r2d2} and ASLFeat~\cite{luo2020aslfeat}, the central idea of our learning technique is to bootstrap the learning process with an existing detector-descriptor pair that is targeted at high confidence matches. Let $A \in \mathbb{R}^{H \times W}$ be the anchor image, which is an image from our training set. On image $A$, we apply two combinations of random homography and thin-plate-spline warp (TPS)~\cite{donato2002approximate}, $g$ and $g'$, to generate images $B$ and $B'$, respectively. The TPS was purposefully chosen since non-rigid deformations can be efficiently parameterized via TPS wrappings. Then, using a base detector, we find $k$ salient keypoints for images $A$, $B$, and $B'$. We use $k = 0.02 \times H \times W$ in our trials, which yields a significant amount of keypoints at all regions in the image. This number of keypoints is an upper limit for the ground truth generation, as we filter a great portion of them later in the pipeline.

\begin{figure}[t!]
%\vspace{-10pt}
  \centering\includegraphics[width=0.99\linewidth]{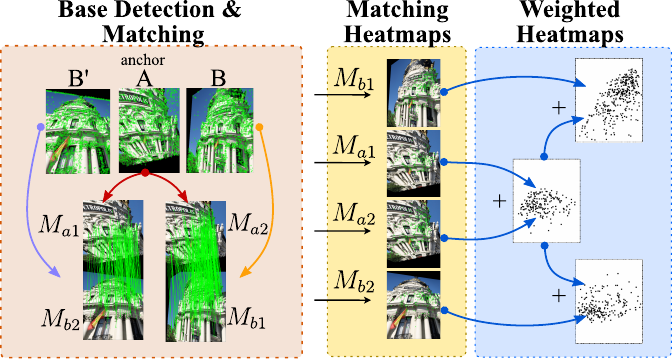}
    \vspace*{-0.7cm}
    \caption{\textbf{Training setting.} Our training comprises three steps: i) We detect the keypoints using a base detector for images A, B and B' (an anchor, and two transformed versions of the anchor with random homography and non-rigid image deformations);
and then, we extract the descriptors on the detected keypoints to find correspondences with nearest neighbor search; 
ii) Using the correct matches, we build a Matching Heatmap (MH) from the locations of correct matches for each input image;
iii) MH weighting done based on keypoint quality, \ie, true match repeatability.}
    \label{fig:gt_generation}
    \vspace{-10pt}
\end{figure}

After selecting the salient pixels, we extract descriptors for each keypoint location and compare the descriptors of images $A$ and $B$, and the descriptors of images $A$ and $B'$. Note that $g$ and $g'$ can be used to determine the locations of the correct matches. When the descriptors pass the nearest neighbor distance ratio tests and is a correct match, its keypoint location $(x, y)$ is added to the set $C_i$, where $i$ is the index of the keypoint for image $A$, $B$, or $B'$. The tolerance used to consider a match correct is set to $3$ pixels, and the nearest neighbor distance ratio is set to $0.8$. Using the position of the correct descriptors' matches as our training data, we train our model to recognize reliable keypoint locations. The map created using true matches is denoted as \textit{Matching Heatmap} (MH). Figure~\ref{fig:gt_generation} shows a summary of this procedure.

Let the MH of $A$ be $M_{a1}, M_{a2}$ and the MH of $B$ be $M_{b1}, M_{b2}$ having values in the range $[0, 1]$, where value $0$ denotes regions with low matching confidence and $1$ denotes regions with high matching confidence (Figure~\ref{fig:gt_generation}). The size of the MH is the same as the input image. If the position $(x,y)$ is in the set $C_i$, we set the MH value to $1$. The final step combines the MH from all pairwise matches such that the areas in the map where descriptors were correctly matched on both match attempts, \ie, matches of image $A$ with $B$ and $A$ with $B'$, are given more weight. The final MH for image $A$ is therefore $M_a = (M_{a1} + M_{a2})/2$. A similar approach with three degrees of weights is applied to images $B$ and $B'$. Regarding image $B$, we have descriptors that are accurate in both image pairs, accurate in the match between $B$ and $A$, and  accurate in the match between $B'$ and $A$. On the MH of $B$, the latter is likewise depicted, albeit with less weight. For $B'$, the same reasoning is used. As a result, the global MHs for $B$ and $B'$ are as follows: $B$: $M_b = (g(M_{a}) + M_{b1})/2$ and $B'$: $M_{b'} = (g'(M_{a}) + M_{b2})/2$. At the end of this operation, we utilize images $B$ and $B'$ and their respective matching heatmaps to train the network. A $3 \times 3$ Gaussian kernel with $\sigma = 1.5$ is applied to each binary MH for it to peak at the ground truth keypoint coordinates. This makes it simpler for the CNN model to learn the global MH. We chose this kernel size to maintain a small activation neighborhood while maintaining smoothness in the heatmaps.

Finally, each image's high-confidence locations to be matched are present on the MH. As we select only repetitive keypoints by choosing the right matches, we also impose our model to learn repeatable points. Siamese training~\cite{bromley1993signature} is used to boost similarities between the score map and the MH of the anchor image and its transformations to further reinforce the repeatability of the detected keypoints. These techniques increase the detector's repeatability when subject to geometric changes. It is important to highlight that our strategy is agnostic to the base detector and descriptor used to generate the ground truth of correspondences. In the experiments, we demonstrate how the suggested detection strategy can enhance the matching ability of three recent descriptors.

\subsection{Loss function}
Due to the uneven distribution of positive and negative pixels in the MH,  using the entire map  for training would bias the model's predictions in favor of maps with generally very low scores. Therefore for each positive keypoint match in the training, we randomly choose a certain number of negative keypoint instances to address this issue. We uniformly sample $n$ negative examples and back-propagate $2n$ examples, where $n$ is the number of positive keypoints in each image. This strategy is expressed as a binary pixel-wise mask $F$ with the value $1$ at the coordinates of the selected pixels and the value $0$ otherwise. We define $S' = S \times F$ given an image $I$, its respective MH $M$, and the model output score map $S$. Finally, we apply the cosine similarity to optimize the similarity between the MH $M$ and the estimated score map image from the network $S'$: 
$$\mathcal{L}_{cossim}(I) = 1 - cossim \left(S', M \right).$$
We chose the cosine similarity to improve the keypoint localization, as this loss focuses more on assuring that the high responses between the ground truth and the predicted map are in the same location rather than on the magnitude of the values between the score maps. Also, to optimize the peaks values, we take the following L2 loss into account:
\begin{equation}
    \mathcal{L}_{simple}(I) = \frac{1}{2n}\sum_{i=1}^{H\cdot W}\left(S'_i - M_i\right)^2.
    \label{eq:loss2}
\end{equation} 
We also want the regressed map to peak at the keypoint locations. Therefore, we use a third loss term to force local peakiness in the score map to achieve even faster convergence. Taking into consideration a collection of non-overlapping $N\times N$ patches $\mathcal{P}=\{p\}$ originated from a regular grid within the image $I$, where there is at least one non-zero pixel at the corresponding location of the patch on M, the peakiness loss term of the score map is defined as:
\begin{equation}
    \mathcal{L}_{peak}(I) = 1 - \frac{1}{|\mathcal{P}|}\sum_{p \in \mathcal{P}}\left(\max_{(i, j) \in p} S_{i, j} - \underset{(i,j) \in p}{\text{mean}} S_{i,j}\right).
    \label{eq:loss3}
\end{equation}
While modeling the loss, we observed that if $N$ is too large, the network will produce fewer keypoints and a smoother heatmap. This behavior is not desirable for pixel-level accurate matching. However, if $N$ is too small, it will produce a high number of low-quality keypoints. The weighted sum of \textit{cossim}, L2, and peak losses yields the final detector training loss $\mathcal{L}$:
%The final loss $\mathcal{L}$ is given by the weighted sum of the \textit{cossim}, L2 and peak losses:
\begin{equation}
    \mathcal{L}(I) = \lambda_1 \mathcal{L}_{cossim}(I) + \lambda_2 \mathcal{L}_{simple}(I) + \lambda_3 \mathcal{L}_{peak}(I).
    \label{eq:loss_final}
\end{equation}
\subsection{Implementation details}
The learned detector comprises a $4$-level deep U-net~\cite{ronneberger2015unet} with a final sigmoid activation function in the output layer. We included $3 \times 3$ convolution blocks with batch normalization and ReLU activations in each layer. Grid-search was used to determine the weights $\lambda_1 =3.0 $, $\lambda_2=1.0$, and $\lambda_3=0.3$ which provided the best results. Even though the experiments are composed of real images with real deformations, our network was only trained using artificially generated deformations. To apply the non-rigid deformations and homography, we employ the code from \cite{potje2021extracting} to generate the simulated deformations. 

The training process used $10K$ images, resulting in 5K pairs of images with a resolution of $400 \times 300$. We applied various random photometric transformations in each image. With an initial learning rate of $0.006$, we optimize the network using Adam, scaling it by $0.9$ every $500$ steps for $7$ epochs. We used a batch size of $12$ images with at least $32$ peaks in their MH. We trained the model on roughly $1.5M$ ground truth keypoints for the ASLFeat descriptor. 
% We implemented the peaky loss by using Max and Average Pooling operations with kernel size $N = 5$.
We implemented the peakness loss by using Max and Average Pooling operations. The window size used in the peakiness term is $N = 5$, which is defined by the kernel size used in the pooling operations. In all experiments, we applied non-maximum suppression (NMS) with a $5\times5$ pixel window size. Additionally, we post-process the keypoints using edge reduction techniques like SIFT~\cite{lowe2004sift} (with a threshold of $10$). The top $k$ keypoints having the largest scores, where $k \in \mathbb{Z}^+$ with respect to detection scores are maintained, while those with detection values below $0.2$ are filtered out. The average time to detect $1{,}024$ keypoints in a $600\times900$ image is 869ms, using 2.5Gb of GPU in an NVIDIA Titan XP.

\begin{figure}[t!]
\vspace{-10pt}
  \centering
    \includegraphics[width=.9\linewidth]{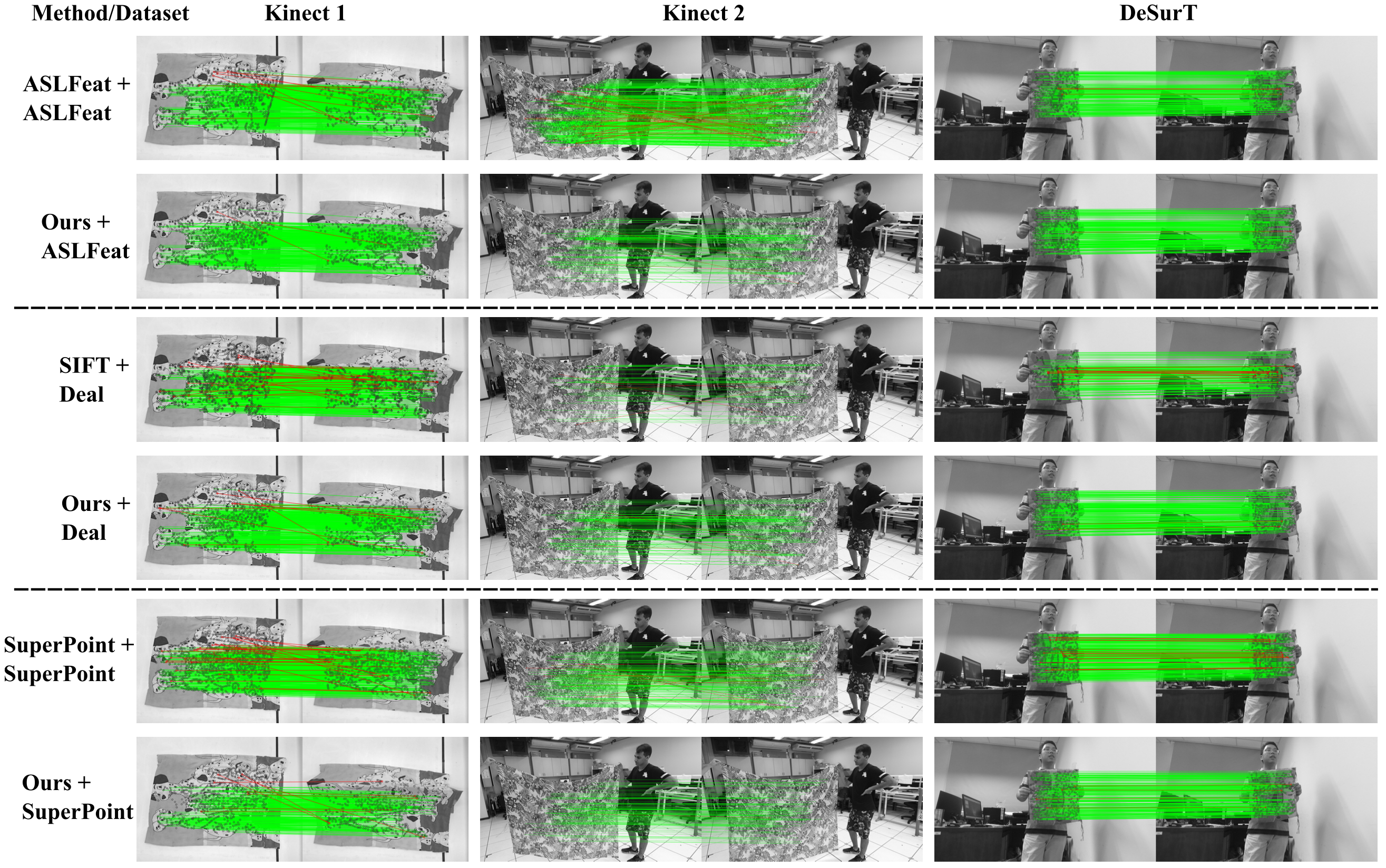}
    \vspace*{-0.4cm}
    \caption{\textbf{Qualitative results on real non-rigid matching examples.} The green lines show correct correspondences, while red lines depict wrong correspondences. Our keypoint detector improves matching accuracy of the DEAL, ASLFeat, and SuperPoint descriptors with respect to their traditional detectors, while maintaining a similar number of total matches.}
    \label{fig:qualitative}
    % \vspace{-10pt}
    \vspace{-0.2cm}
\end{figure}

\section{Experiments and results}
\label{experiments}
The training and validation datasets were generated from random images from SfM datasets~\cite{wilson_eccv2014_1dsfm}. We assess our detector using a variety of publicly accessible datasets with deformable objects under various viewing circumstances, including lighting, perspective, and deformation. We chose the two datasets recently issued by GeoBit and DeSurT~\cite{wang2019deformable}. They contain color images of $11$ deforming real-world objects, with a total of $770$ pairs of images with shapes between $960\times540$ and $640\times480$, and ground-truth correspondences collection follows the protocol of \cite{jin2021image}. Our training setting considers a base detector and descriptor. Given the large range of detectors and descriptors available, we start using ASLFeat~\cite{luo2020aslfeat} because its architecture features deformable convolutional kernels and is among the state-of-the-art methods for the detection-description task. The deformable kernels aim to develop dynamic receptive fields to accommodate the capabilities of modeling geometric variations. We also select the DEAL~\cite{potje2021extracting} descriptor, which is non-rigid deformation invariant. 
At last, to show the robustness of our training setting, we chose the SuperPoint detector-descriptor due to its success on traditional rigid matching tasks.

\begin{table*}[h!]
    \centering
    \caption{Detector matching performance comparison. Best in bold and second-best underlined. The higher the value, the better. Our detector provides keypoints on images with non-rigid deformations that enhances the matching results.}
    \resizebox{\textwidth}{!}{%
    \begin{tabular}{lllllllllll}
    \toprule
    \multicolumn{11}{c}{Dataset $770$ pairs total - MS / MMA@3 pixels} \\ 
    
    \midrule
    
    \multicolumn{1}{c}{\begin{tabular}[c]{@{}c@{}}\textbf{Detector} \\ \textbf{+} \\ \textbf{ASLFeat}\end{tabular}} & \multicolumn{1}{c}{Kinect1} & \multicolumn{1}{c}{Kinect2} & \multicolumn{1}{c}{DeSurT} & \multicolumn{1}{c}{Mean} & \multicolumn{1}{c}{} & \multicolumn{1}{c}{\begin{tabular}[c]{@{}c@{}}\textbf{Detector} \\ \textbf{+} \\ \textbf{DEAL}\end{tabular}} & \multicolumn{1}{c}{Kinect1} & \multicolumn{1}{c}{Kinect2} & \multicolumn{1}{c}{DeSurT} & \multicolumn{1}{c}{Mean} \\ \cline{1-5} \cline{7-11} 
    SIFT            & 0.35 / 0.77 & 0.37 / 0.85 & 0.26 / 0.63 & 0.33 / 0.75 & & SIFT        & 0.33 / 0.68 & 0.38 / \textbf{0.85} & 0.27 / \textbf{0.63} & 0.33 / \underline{0.72} \\
    FAST            & 0.43 / 0.69 & 0.53 / 0.85 & 0.33 / 0.56 & \underline{0.43} / 0.70 & & FAST        & 0.36 / 0.58 & \textbf{0.51} / 0.81 & \underline{0.29} / 0.49 & 0.39 / 0.63 \\
    AKAZE           & 0.39 / 0.66 & 0.49 / 0.76 & 0.26 / 0.48 & 0.40 / 0.66 & & AKAZE       & 0.38 / 0.65 & 0.47 / 0.74 & 0.23 / 0.42 & 0.36 / 0.60 \\
    Keynet          & 0.31 / 0.65 & 0.35 / 0.62 & 0.24 / 0.51 & 0.30 / 0.59 & & Keynet      & 0.27 / 0.58 & 0.34 / 0.59 & 0.22 / 0.45 & 0.28 / 0.54 \\
    ASLFeat         & 0.31 / 0.58 & 0.39 / 0.69 & 0.28 / 0.53 & 0.33 / 0.60 & & ASLFeat     & 0.31 / 0.66 & 0.40 / 0.73 & 0.25 / 0.54 & 0.32 / 0.64 \\
    R2D2            & 0.20 / 0.51 & 0.24 / 0.57 & 0.15 / 0.40 & 0.20 / 0.49 & & R2D2        & 0.16 / 0.41 & 0.20 / 0.48 & 0.12 / 0.33 & 0.16 / 0.41 \\
    ORB             & 0.17 / 0.37 & 0.26 / 0.57 & 0.14 / 0.34 & 0.19 / 0.43 & & ORB         & 0.03 / 0.07 & 0.14 / 0.34 & 0.06 / 0.15 & 0.08 / 0.19 \\
    D2NET           & 0.32 / 0.83 & 0.37 / \textbf{0.89} & 0.18 / \textbf{0.68} & 0.29 / \underline{0.80} & & D2NET       & 0.28 / \underline{0.72} & 0.34 / \underline{0.84} & 0.16 / 0.57 & 0.26 / 0.71 \\
    DISK            & \textbf{0.57} / \underline{0.85} & 0.51 / \underline{0.88} & \textbf{0.40} / \underline{0.66} & \textbf{0.49} / \underline{0.80} & & DISK        & \textbf{0.48} / 0.71 & 0.46 / 0.80 & \textbf{0.35} / 0.55 & \textbf{0.43} / 0.69 \\
    ALIKE           & 0.45 / 0.68 & \underline{0.53} / 0.84 & 0.31 / 0.48 & \underline{0.43} / 0.67 & & ALIKE       & 0.4  / 0.61 & \underline{0.49} / 0.78 & \underline{0.29} / 0.44 & 0.39 / 0.61 \\
    SuperPoint      & 0.46 / 0.77 & \textbf{0.56} / \textbf{0.89} & \underline{0.34} / 0.61 & 0.42 / 0.76 & & SuperPoint  & 0.34 / 0.57 & 0.47 / 0.75 & 0.27 / 0.48 & 0.36 / 0.60 \\ \hdashline \noalign{\vskip 0.5ex} 
    Ours            & \underline{0.49} / \textbf{0.86} & 0.48 / \textbf{0.89} & 0.31 / \underline{0.66} & \underline{0.43} / \textbf{0.81} & & Ours        & \underline{0.45} / \textbf{0.79} & 0.46 / \textbf{0.85} & 0.28 / \underline{0.59} & \underline{0.40} / \textbf{0.74} \\
    \bottomrule
    \end{tabular}%
    }
    
    % }
    \label{tab:results_match}
    \vspace{-0.4cm}
    \end{table*}
    
\subsection{Metrics and baselines}
\label{experiments:metrics}

In our experiments, we use the Mean Matching Accuracy (MMA)~\cite{revaud2019r2d2} that computes the ratio of correct matches to possible matches; and the Matching Score (MS)~\cite{revaud2019r2d2}. We chose these metrics since the ultimate objective of our feature detection is to maximize the number of correct feature matches. We also employ keypoint Repeatability Rate (RR), a popular keypoint metric that computes the ratio of potential matches to the minimum number of keypoints in the shared view with a $3$ pixel error threshold. We perform comparisons to eleven widely adopted detectors across three descriptors. We take into account three handcrafted detectors, SIFT~\cite{lowe2004sift}, ORB, FAST~\cite{rosten2006fast} and AKAZE~\cite{alcantarilla2012kaze}, which provide stable keypoints and are still regarded as good baselines by a recent study~\cite{jin2021image}; and seven state-of-the-art learning-based frameworks: Keynet~\cite{laguna2022keynet}, ASLFeat~\cite{luo2020aslfeat}, R2D2~\cite{revaud2019r2d2}, D2Net~\cite{dusmanu2019d2net}, DISK~\cite{tyszkiewicz2020disk}, ALIKE~\cite{Zhao2022ALIKE} and SuperPoint~\cite{DeTone_2018_CVPR_Workshops}. Considering the discussion of Mikolajczyk \etal \cite{mikolajczyk2005comparison}, the metrics require the methods to detect enough keypoints for the stability of the results. For that reason, on each image, each detector is configured to find $1,024$ keypoints. Methods like SEKD and GLAM could not maintain a large number of detections and, for that reason, are not included in Table \ref{tab:results_match}
\subsection{Results on sequences of deformable objects}
\label{experiments:results}

We evaluate the metrics achieved by the detectors using existing descriptors. For this, we chose DEAL~\cite{potje2021extracting}, ASLFeat~\cite{luo2020aslfeat}, and SuperPoint~\cite{DeTone_2018_CVPR_Workshops} as descriptors and re-trained three detectors to match each of the three considered descriptors. As can be observed in Table~\ref{tab:results_match}, in both MS and MMA metrics, our keypoints combined with DEAL descriptors perform better on average than all detector-DEAL combinations. It is also clear that our trained detector achieves the best MMA across all datasets for the ASLFeat descriptor. It is important to note that when the ASLFeat's detector is replaced by ours, the average MMA score jumps from $0.60$ to $0.80$ ($20$ p.p.), and our approach is significantly different from the second-best MMA method by $4$ p.p (SuperPoint-ASLFeat). When compared to the SIFT detector used to train the DEAL descriptor, our detector consistently gets the highest and second-best MS and MMA scores, improving, on average, by roughly $7$ and $2$ points for MS and MMA respectively. Matching examples of our detector using various descriptors are shown in Figure~\ref{fig:qualitative}. Similar to the SIFT and ASLFeat detectors, our technique can provide evenly spaced matches in the image, but with higher accuracy.

Aside from MMA and MS, we also evaluated the repeatability rate, where FAST has the best RR of any approach, averaging $0.59$ and followed by SuperPoint at $0.57$. AKAZE, ASLFeat, and our approach achieve an RR of $0.50$. SIFT has the lowest RR metric of any detector with $0.43$. These results demonstrate that our detector achieves a competitive RR while improving the MMA and MS matching metrics. It is noteworthy that, while having a high RR, the ASLFeat detector has a smaller MS and MMA, as shown in Table~\ref{tab:results_match}. Moreover, a high RR does not necessarily translate into strong matching results. We also extended our experiments to retrain our detector using the widely adopted learned detector-descriptor SuperPoint. In Table~\ref{tab:results_match_sp}, we compare the results of all the previous baseline detectors in conjunction with SuperPoint descriptors. The "Ours [SP]" line shows the results of training on SuperPoint keypoints and descriptors. Even though SuperPoint has a joint detector-descriptor learning method, we were able to better filter the detection in a way that we have better matches at the cost of less repeatable keypoints. We highlight that the SuperPoint training already includes keypoint augmentation via homography to improve repeatability, thus it would be very challenging to improve upon that without losing match accuracy. We gain $6$ p.p. in MMA, showing that our keypoints are still competitive for the descriptor.
\begin{table}
	\vspace*{-0.5cm}	
	\caption{Detector matching performance comparison. Best in bold and second-best underlined.}
	\resizebox{.48\textwidth}{!}{%
		
		\begin{tabular}{lcccc}
			\toprule
			\multicolumn{5}{c}{Dataset 770 pairs total - MS / MMA@3}    \\ \hline
			\multicolumn{1}{c}{\begin{tabular}[c]{@{}c@{}}\textbf{Detector} \\ \textbf{+}\\ \textbf{SuperPoint}\end{tabular}} & Kinect1     & Kinect2     & DeSurT      & Mean        \\ \midrule
			SIFT            & 0.34 / 0.74 & 0.36 / 0.85 & 0.25 / \underline{0.63} & 0.32 / 0.74 \\
			FAST            & 0.38 / 0.79 & 0.39 / 0.55 & 0.30 / 0.48 & 0.36 / 0.61 \\
			AKAZE           & 0.41 / 0.69 & 0.47 / 0.69 & 0.28 / 0.59 & 0.39 / 0.67 \\
			Keynet          & 0.28 / 0.59 & 0.37 / 0.64 & 0.24 / 0.51 & 0.30 / 0.58 \\
			ASLFeat         & \underline{0.48} / \textbf{0.84} & \textbf{0.53} / 0.78 & \textbf{0.34} / 0.62 & \underline{0.45} / 0.75 \\
			R2D2            & 0.18 / 0.48 & 0.22 / 0.54 & 0.14 / 0.37 & 0.18 / 0.46 \\
			ORB             & 0.14 / 0.31 & 0.23 / 0.51 & 0.13 / 0.30 & 0.17 / 0.37 \\
			D2NET           & 0.31 / \underline{0.80} & 0.35 / \underline{0.87} & 0.18 / \textbf{0.67} & 0.28 / \textbf{0.78} \\
			DISK            & \textbf{0.53} / \underline{0.80} & 0.49 / 0.86 & 0.39 / \underline{0.63} & \textbf{0.47} / \underline{0.76} \\
			ALIKE           & 0.41 / 0.62 & \underline{0.50} / 0.79 & 0.29 / 0.45 & 0.40 / 0.60 \\
			SuperPoint      & 0.43 / 0.73 & \textbf{0.53} / 0.85 & \underline{0.33} / 0.59 & 0.43 / 0.72 \\ \hdashline \noalign{\vskip 0.5ex}
			Ours {[}SP{]}   & 0.33 / 0.78 & 0.45 / \textbf{0.88} & 0.21 / \textbf{0.67} & 0.33 / \textbf{0.78} \\ \bottomrule
		\end{tabular}
	}
	\label{tab:results_match_sp}
	\vspace*{-0.5cm}
\end{table}
Tables \ref{tab:results_match} and \ref{tab:results_match_sp} show that the ALIKE descriptor has a small advantage in matching scores in some cases due to the reduced number of keypoints (approximately 100 detected keypoints less on average), but is inferior in matching accuracy for all experiments. Also our approach is superior to DISK in MMA in almost all experiments, while having a competitive performance in Matching Scores. 
Despite SEKD and GLAM not detecting enough instances (averaging 121 and 235, respectively), we computed their MMA, MMS, and RR metrics using the nonrigid descriptor DEAL. For the sake of fairness, we also lowered the number of maximum de- tections of our method to match each competitor. Com- paring Ours versus SEKD, on average, SEKD scored 0.51, 0.43, and 0.80 in RR, MMS, and MMA, respectively, while Our detector scored 0.44, 0.36, and 0.78. Comparing Ours versus GLAM, on average, GLAM scored 0.51, 0.42, 0.79 in RR, MMS, and MMA, respectively, while our detector scored 0.47, 0.40, and 0.80. It is worth mentioning that experiments with a small number of keypoints can result in misleading results as shown by Mikolajczyk \etal \cite{mikolajczyk2005comparison}

\subsection{Results on the application of object retrieval} 
 % Please add the following required packages to your document preamble:
% \usepackage{graphicx}
\begin{table}[t!]
\vspace{-0.25cm}
\centering
\caption{\textbf{Ablation study.} Comparing MS, MMA, and Mean RR for different architecture choices.}
 \resizebox{.5\textwidth}{!}{%
\begin{tabular}{lllll}
\multicolumn{5}{c}{Dataset 770 pairs total - MS / MMA@3 / MRR pixels}            \\ \hline
Our Detector        & Kinect1        & Kinect2        & DeSurT         & Mean           \\ \hline
Enc. DCN            & 0.21/0.67/0.22 & \underline{0.41}/\underline{0.93}/\underline{0.43} & 0.21/0.52/0.37 & 0.27/0.70/0.42 \\
Dec. DCN            & 0.44/0.82/0.48 & 0.37/0.92/0.40 & 0.29/0.64/0.26 & 0.37/0.79/0.30 \\
All DCN             & \underline{0.48}/\textbf{0.92}/\underline{0.52} & 0.3/\textbf{0.94}/0.31  & \textbf{0.36}/\textbf{0.68}/\textbf{0.46} & \underline{0.38}/\textbf{0.84}/\underline{0.43} \\ 
Single Branch       & 0.43/\underline{0.88}/0.48 & 0.26/0.92/0.28 & 0.22/0.6/0.30  & 0.3/\underline{0.82}/0.35 \\ 
Only Homography     & 0.37/0.86/0.43 & 0.24/\underline{0.93}/0.25 & 0.26/0.64/0.37 & 0.29/0.81/0.35 \\  \hdashline \noalign{\vskip 0.5ex} 
Ours                & \textbf{0.49}/0.86/\textbf{0.55} & \textbf{0.48}/0.89/\textbf{0.54} & \underline{0.31}/\underline{0.66}/\underline{0.42} & \textbf{0.43}/0.80/\textbf{0.51} \\
\bottomrule
\end{tabular}%
}
\label{table:ablation_dcn}
\vspace{-0.25cm}
\end{table}

To further demonstrate the efficacy of our detector in real-world applications, we conducted experiments on a content-based object retrieval task. The goal is to return the top K images corresponding to a given query. We represent each image using a Bag-of-Visual-Words method. The DEAL descriptor~\cite{potje2021extracting} was used to create a visual dictionary for each keypoint, which is then used to generate a global descriptor for each image. We determine the global descriptor for a given query image and employ the K-Nearest Neighbor search to identify the top K closest objects. To assess the effectiveness of the detectors, we use retrieval accuracy@K, \ie, the number of accurate objects recovered in the top K images. We only employ a descriptor that models isometric deformations because the application's database is deformable. The retrieval accuracy for SIFT, ASLFeat, and Ours at $K=20$ achieved $100\%$. The outcomes show that, even though our detector was trained to improve matching tasks, it can effectively select representative keypoints for non-matching challenges.

\subsection{Ablation study and sensitivity analysis}
\label{experiments:ablation}
\noindent\textbf{Siamese network:} We trained our model using two configurations for ablation studies: (i) a Siamese network scheme and (ii) a conventional network training scheme, \ie,  employing a single branch. According to the experiments, a Siamese scheme (i) aids the model's ability to learn repeating keypoints. Using the Siamese method, the RR went up from $0.35$ to $0.51$, the MS from $0.39$ to $0.43$, and the MMA went down from $0.81$ to $0.80$; nevertheless, the inliers dramatically increased from $150$ to $170$.

\noindent\textbf{Deformable convolutions:} We also evaluated the contribution of adopting Deformable Convolutional layers (DCNs) in the network model of our detector with four configurations: (i) using DCNs at the decoder (dec.), (ii) using DCNs at the encoder (enc.), (iii) using DCNs in both the encoder and decoder, and (iv) using conventional convolution layers. We can see in the results of Table~\ref{table:ablation_dcn} the adopted full CNN model achieved, on average, the best matching score and repeatability rate while maintaining a consistently high, and second best, matching accuracy. 
Since DCN layers have considerably more learnable parameters, they introduce a bigger domain gap between the synthetic training data and the real-world deformable objects on the benchmark datasets.

\noindent\textbf{TPS deformations:} We trained our model with only homography changes to evaluate the effect of using TPS on top of homography to generate the dataset. As shown in table \ref{table:ablation_dcn}, the use of TPS deformations improved every metric on all datasets, showing that this transformation is much more representative of real nonrigid objects than simply homography.

\begin{table}[t!]
    \centering
    \vspace*{-0.3cm}
    \caption{Ablation study. Comparing the average MS, MMA, and Mean RR across the nonrigid evaluation dataset  for different loss combinations.}
    \label{tab:loss_ablation}
     \resizebox{.46\textwidth}{!}{%

    \begin{tabular}{llll}
        \hline
        \textbf{Loss Combination}            & RR            & MMS           & MMA@3         \\ \hline
        (i) \ \  $L_{cossim} + L_{simple} + L_{peak}$ & 0.50          & \textbf{0.43} & 0.81          \\
        (ii) \ $L_{cossim} + L_{simple}$            & 0.48          & 0.39          & 0.80          \\
        (iii) $L_{cossim} + L_{peak}$              & 0.34          & 0.29          & 0.80          \\
        (iv) $L_{simple} + L_{peak}$              & 0.25          & 0.19          & 0.74          \\
        (v) \  $L_{cossim}$                         & 0.42          & 0.37          & 0.81          \\
        (vi) $L_{simple}$                         & \textbf{0.51} & \textbf{0.43} & \textbf{0.82} \\ 
        \hline
    	\end{tabular}
     }
    \vspace{-0.25cm}

\end{table} 

\begin{figure}[t!]
	%\vspace{-10pt}
	\centering\includegraphics[width=0.99\linewidth]{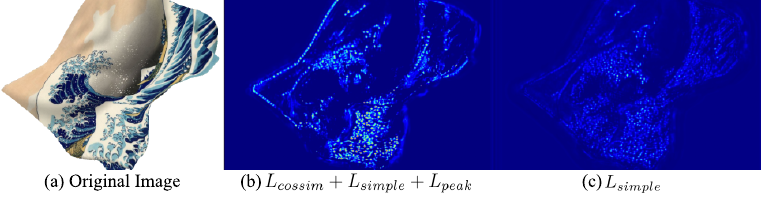}
	\vspace*{-0.7cm}
	\caption{\textbf{Loss comparison.} Looking closer at the score map prediction between the configurations (b) and (c), we can see that combining the three losses (b) results in much more confident keypoints with less ambiguous areas.
	}
	\label{fig:loss_experiment}
	\vspace{-10pt}
\end{figure}

\noindent\textbf{Loss combinations:} At last, we assess six setups to gauge how well each of our proposed loss components (Equation~\ref{eq:loss_final}) contributes to and supports our implementation choices, as shown in table \ref{tab:loss_ablation}. Our findings demonstrate that setup (i) is the most effective, with the repeatability rate at $0.5$,  the best MS at $0.43$, and MMA@3 at $0.81$. Despite (vi) having slightly better values, it results in much less confident keypoints, resulting in an ambiguous score map with almost no peaks, as shown in Figure \ref{fig:loss_experiment}. For this reason we choose setup (i).
To further justify the weighting step choices in our training framework, we experimented by applying equal weights based on repeatable matching. This will set all the peaks in the MHs to a constant value of $1.0$. For this equal weights strategy, we achieved RR and MS values of $0.45$ and $0.37$, which are much lower than what we achieve when utilizing the proposed weighted Matching Heatmap strategy ($0.50$ and $0.43$ for RR and MS, respectively).

{\noindent\textbf{Limitations:}
As stated in Section \ref{sec:methodology}, our detection strategy learns to improve the types of keypoints provided by the base detector and conditions them to work better with the descriptor used in the training for matching non-rigid surfaces. This behavior is not uncommon, and it can be observed when a detector is trained with the assistance, or at the same time, of a specific descriptor, like R2D2 and DISK. But this modeling also implies that our detector can incorporate limitations from the description method, e.g., if ASLFeat fails to match poorly-textured regions, our detector will also avoid these areas.

\section{Conclusion}
\label{conclusion}\vspace*{-0.1cm}
We present a novel method for locating confident keypoints on images having non-rigid deformations, focused on improving matching results. For designing a detector that learns the likelihood of proper matching for a given descriptor, we devised a method for training a CNN using non-rigidly deformed images. The experimental results demonstrate that our approach successfully matched and detected keypoints on non-rigid deformation datasets with state-of-the-art accuracy.
We observed that learning the types of keypoints that remain robust to matching goes beyond keeping the base detector predictions, as our method could maintain the number of detections and improve the keypoint selection.
The experimental analysis also indicates that a successful detector requires more than just repeatability. Our learned detector improved matching for three different state-of-the-art descriptors compared to a wide variety of competitors. We also demonstrate the effectiveness of our detector in a practical setting, indicating the potential of improving keypoint detection as a research area for enhancing performance in practical tasks.
Future works can easily extend our approach to tackle not only challenges in deformable object matching but a wider class of problems involving reliable keypoint computation.
\vspace*{0.1cm}
\newline 
\noindent\textbf{Acknowledgments.}
\noindent We would like to thank CAPES, CNPq, FAPEMIG, Google, and Conseil Régional BFC for funding different parts of this work and NVIDIA for the donation of a Titan XP GPU used for this study. This work was also granted access to the HPC resources of IDRIS under the project 2021-AD011013154.

\vspace{-0.3cm}
\bibliographystyle{elsarticle-num}
\bibliography{prl_main}

\end{document}